\documentclass[letterpaper, 10 pt, conference]{ieeeconf}  

\IEEEoverridecommandlockouts 

\usepackage{graphics} 
\usepackage{amsmath,amssymb,amsfonts} 
\usepackage{bm}
\usepackage{caption}	 
\usepackage{subcaption}	 
\usepackage{graphicx}
\usepackage{multirow}
\usepackage{array}
\usepackage{booktabs}
\usepackage[ruled,vlined]{algorithm2e}
\usepackage{xcolor}
\usepackage{gensymb}
\usepackage{svg}
\usepackage{amsmath}
\usepackage{tikz}
\usepackage{makecell}
\svgsetup{
    inkscapepath=i/svg-inkscape/
}
\svgpath{{svg/}}

\usepackage[backend=bibtex,natbib=true,style=ieee,sorting=none,giveninits=true,maxbibnames=99,url=false,doi=false]{biblatex}
\DeclareFieldFormat[inproceedings,article]{title}{#1}

\title{\LARGE \bf
Structure-Aware Radar-Camera Depth Estimation 
}

\author{Fuyi Zhang, Zhu Yu$^{*}$, Chunhao Li, Runmin Zhang, Xiaokai Bai, Zili Zhou, Si-Yuan Cao, Fang Wang, and \\ Hui-Liang Shen$^{*}$, \emph{Senior Member, IEEE}
\thanks{This work was supported in part by the National Key Research and Development Program of China under grant 2023YFB3209800 and in part by the National Natural Science Foundation of China under grant 62301484.}
\thanks{
Fuyi Zhang, Zhu Yu, Chunhao Li, Runming Zhang, Xiaokai Bai, and Zili Zhou are with the College of Information Science and Electronic Engineering, Zhejiang University, Hangzhou 310027, China (e-mail: fuyizhang@zju.edu.cn, yu\_zhu@zju.edu.cn, li\_chunhao@zju.edu.cn, runmin\_zhang@zju.edu.cn, xiaokaibai@zju.edu.cn, zhou\_zili@zju.edu.cn).}
\thanks{
Si-Yuan Cao is with the Ningbo Innovation Center, Zhejiang University, China (e-mail: cao\_siyuan@zju.edu.cn).}
\thanks{
Fang Wang is with the School of Information and Electrical Engineering, Hangzhou City University, China, and also with the Hangzhou City University Binjiang Innovation Center, China (e-mail: wangf@zucc.edu.cn).}
\thanks{
Hui-Liang Shen is with the College of Information Science and Electronic Engineering, Zhejiang University, and also with the Key Laboratory of Collaborative Sensing and Autonomous Unmanned Systems of Zhejiang Province, China (e-mail: shenhl@zju.edu.cn).}
\thanks{$^*$ Zhu Yu and Hui-Liang Shen are the corresponding authors.}
}

\begin{document}

\maketitle
\thispagestyle{empty}
\pagestyle{empty}

\begin{abstract}
Radar has gained much attention in autonomous driving due to its accessibility and robustness. However, its standalone application for depth perception is constrained by issues of sparsity and noise. Radar-camera depth estimation offers a more promising complementary solution. Despite significant progress, current approaches fail to produce satisfactory dense depth maps, due to the unsatisfactory processing of the sparse and noisy radar data. They constrain the regions of interest for radar points in rigid rectangular regions, which may introduce unexpected errors and confusions. To address these issues, we develop a structure-aware strategy for radar depth enhancement, which provides more targeted regions of interest by leveraging the structural priors of RGB images. Furthermore, we design a Multi-Scale Structure Guided Network to enhance radar features and preserve detailed structures, achieving accurate and structure-detailed dense metric depth estimation. Building on these, we propose a structure-aware radar-camera depth estimation framework, named SA-RCD. Extensive experiments demonstrate that our SA-RCD achieves state-of-the-art performance on the nuScenes dataset. Our code will be available at \url{https://github.com/FreyZhangYeh/SA-RCD}.
\end{abstract}
\vspace{-3pt}

\section{Introduction}

Perceiving accurate depth is vital for 3D perception in autonomous driving ~\cite{song2021self,cui2021deep,li2023bevdepth,zhou2022towards,alaba2024emerging,wang2023multi, CGFormer}. Commercially available depth sensors like LiDAR and radar, can directly capture metric depth from the environment. While LiDAR~\cite{cspn, cspn2, nlspn, yan2022rignet, tpvd, pointdc} produces higher-precision and higher-density point cloud data compared to radar, it comes with higher costs, larger size, and is more sensitive to environmental factors such as rain, fog, and snow \cite{wang2023multi,alaba2024emerging}. In contrast, radar has recently attracted increasing attention due to its ease of deployment and robustness to adverse lighting and weather conditions \cite{wang2023multi,alaba2024emerging,zhou2022towards,han23064d}.

However, radar faces inherent challenges, such as sparsity, noisy detections, and poor semantic information, due to its imaging mechanism and multi-path propagation \cite{wang2023multi,alaba2024emerging,zhou2022towards,RC-PDA,R4Dyn,Radarnet}. Consequently, various researches integrate radar alongside cameras for depth estimation \cite{RC-PDA,long2021full,LinDepth,RegressionDepth,Radarnet,R4Dyn,RCDPT} to overcome the aforementioned limitations, known as radar-camera depth estimation. The complementarity between the depth cues of radar and the scene priors of camera makes it easier to estimate dense and accurate metric depth, which appears to be a promising solution.
\begin{figure}[t]
     \centering 
     \begin{subfigure}[t]{\linewidth}
      \centering
    \includegraphics[width=0.9\linewidth]{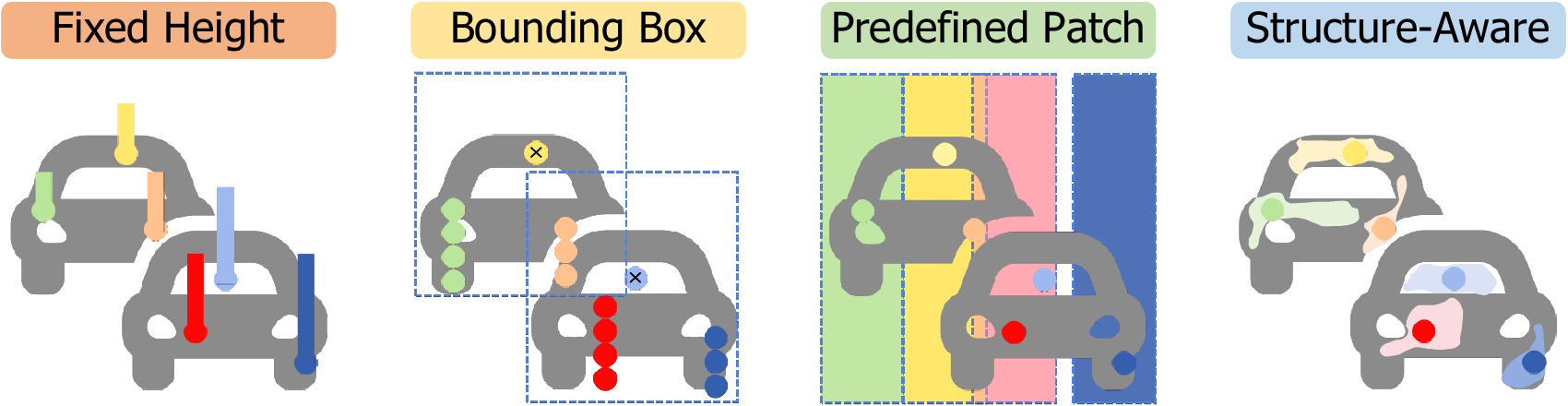}
    \vspace{-5pt}
    \captionsetup{font=footnotesize}
    \caption{ Illustration of hypothesized ROIs for radar points.}
    \label{fig:head1}
     \end{subfigure}
     \begin{subfigure}[t]{\linewidth}
      \centering
    \includegraphics[width=0.9\linewidth]{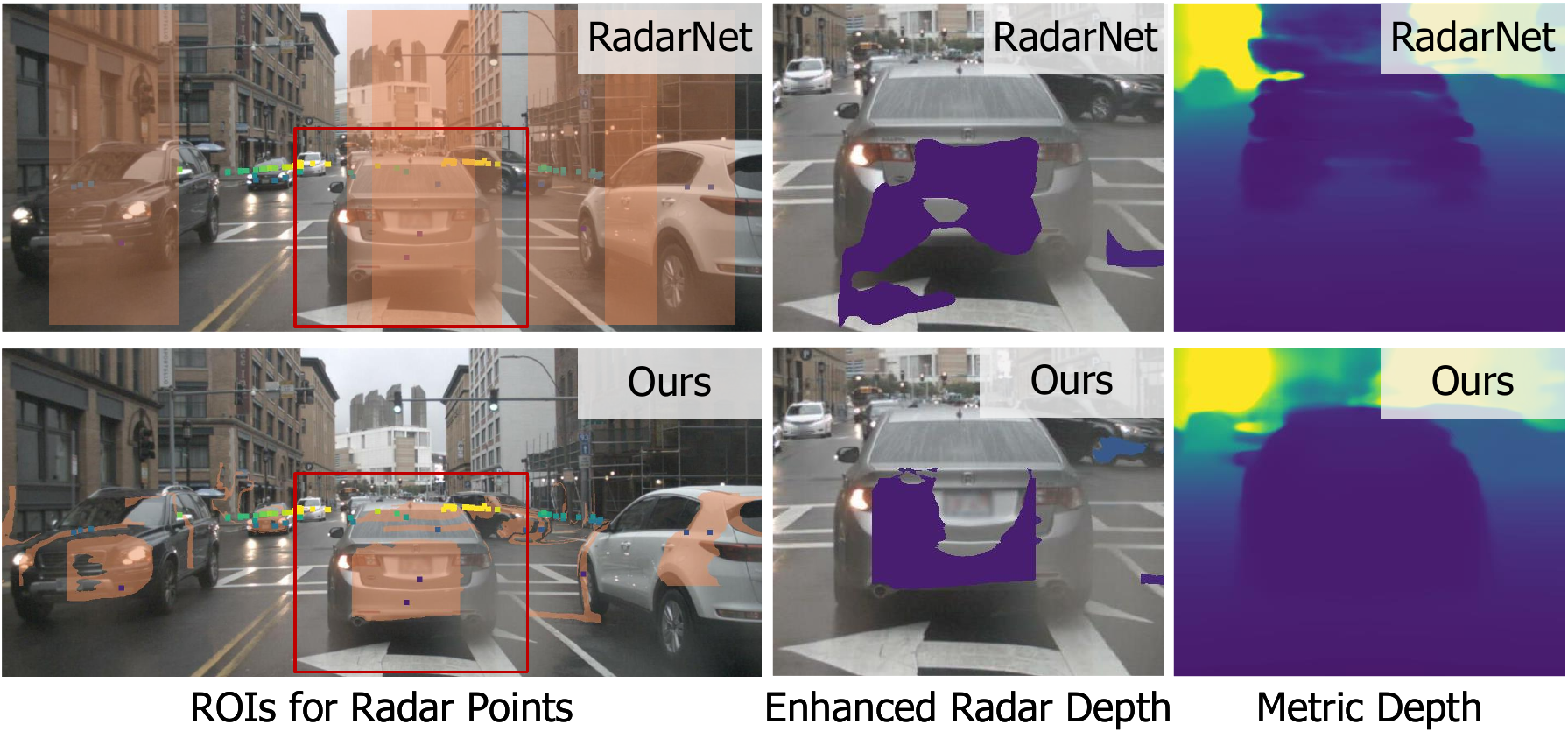}
    \vspace{-5pt}
    \captionsetup{font=footnotesize}
    \caption{An example of ROIs, enhanced radar depth, and metric depth.}
    \label{fig:head2}
     \end{subfigure}
    \vspace{-5pt}
     \captionsetup{font=footnotesize}
     \caption{A comparison of our hypothesized ROIs for radar points and radar depth estimation performance against other approaches. (a) Previous approaches constrain the hypothesized ROIs for radar points to rectangular shapes, such as fixed height \cite{RegressionDepth}, bounding box \cite{R4Dyn} and predefined patch \cite{Radarnet}, while ours employ a structure-aware strategy to yield ROIs adaptively.
     (b) Compared with the previous state-of-the-art approach RadarNet \cite{Radarnet}, our SA-RCD produces more targeted ROIs for radar depth enhancement and estimates more structure-detailed metric depth.}
     \label{fig:cover} 
     \vspace{-18pt} 
\end{figure} 

\begin{figure*}[t]
      \centering
      \includegraphics[width=0.98\textwidth]{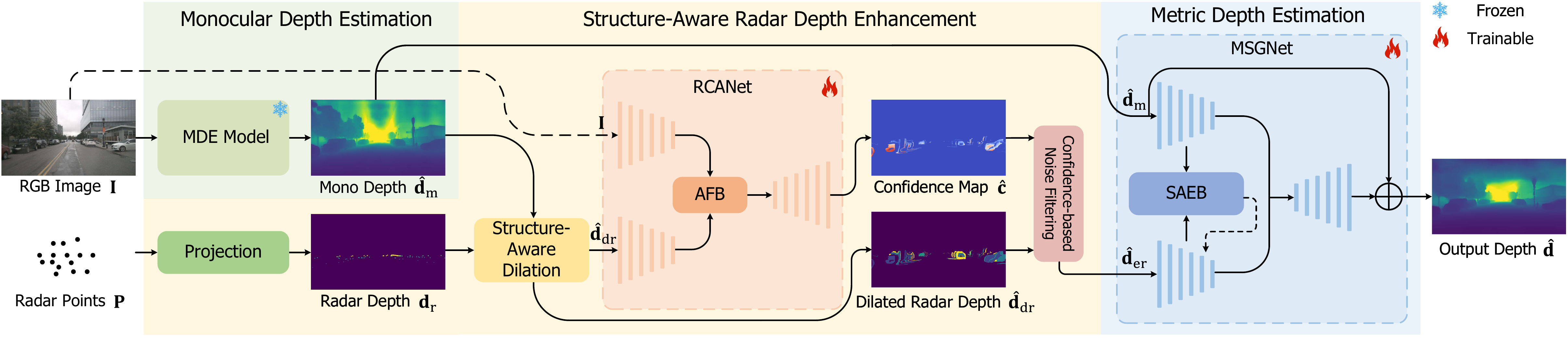}
      \caption{Schematics of the proposed SA-RCD. The framework consists of the monocular depth estimation for capturing structure priors from RGB images, the structure-aware radar depth enhancement for enhancing radar depth in targeted regions, and the metric depth estimation through MSGNet for achieving accurate and structure-detailed metric depth estimation.}
      \label{fig:framework}
      \vspace{-15pt} 
\end{figure*}

Current radar-camera depth estimation methods typically employ a two-stage framework. The first stage aims to enhance radar data, while the second stage estimates dense metric depth using the enhanced radar data and RGB images. A common radar data enhancement strategy in the first stage is to merge radar point clouds from adjacent frames into the current frame for densification \cite{RC-PDA,LinDepth,RegressionDepth,R4Dyn,RCDPT}. However, this may be impractical in real-world applications due to latency and unavailability \cite{Radarnet}. Alternative approaches extend radar depth along the height dimension or within bounding boxes \cite{RegressionDepth,R4Dyn}. More advanced approaches attempt to model the uncertainty of radar points by learning radar-camera pixel association within predefined patches \cite{RC-PDA,Radarnet,li2024radarcam}. However, all of these approaches constrain the regions of interest (ROIs) \cite{Radarnet} for radar points to rigid rectangular shapes, without accounting for the structure priors inherent in the scene, as shown in Fig. \ref{fig:cover}. This may introduce additional erroneous measurements and confuse the learning of radar-camera pixel association. Then, the unsatisfactory enhanced radar data is fed into the second stage, undermining accurate and structure-detailed metric depth estimation.

To tackle the above issues, we aim to achieve structure-aware radar-camera depth estimation by leveraging the structure priors in RGB images. The recently emerging zero-shot monocular depth estimation (MDE) methods \cite{MiDaS,Dany,Zoedepth,ZeroDepth} provide a strong support for our idea. They have demonstrated great potential in extracting valuable structural priors related to depth features from unseen images. Although these models struggle to estimate accurate metric depth, they produce structure-detailed monocular depth with fine relative geometric relationship. Considering this, we propose a framework consisting of three stages for radar-camera depth estimation, named SA-RCD, as illustrated in Fig. \ref{fig:framework}. In the first stage, we leverage a MDE model to effectively capture structure priors from RGB images, generating structure-detailed monocular depth. In the second stage, we develop a structure-aware strategy to enhance radar depth in more targeted ROIs. Instead of being constrained in rigid rectangular shapes, these ROIs are grown adaptively centered on radar pixels, guided by the distribution of monocular depth, as shown in Fig. \ref{fig:cover}. In the third stage, we fuse the enhanced radar depth and monocular depth to estimate dense metric depth through a well-designed Multi-scale Structure Guided Network (MSGNet). Specially, we employ residual-oriented learning to preserve fine structure details in monocular depth, and further integrate a Structure-Aware Enhancement Block (SAEB) to enhance radar features at multiple scales. In summary, the main contributions of this work are as follows:
\begin{itemize}
    \item We propose SA-RCD, a novel structure-aware radar-camera depth estimation framework consisting of  three stages: monocular depth estimation, structure-aware radar depth enhancement, and metric depth estimation. Extensive experiments demonstrate that SA-RCD achieves the state-of-the-art performance on the nuScenes dataset \cite{caesar2020nuscenes}.
    \item We develop a structure-aware strategy to enhance radar depth. It leverages the structure priors of RGB images to guide the generation of ROIs for radar points, providing more targeted regions for radar depth enhancement.
    \item We design a Multi-scale Structure Guided Network (MSGNet) to estimate dense metric depth. It employs residual oriented-learning for structure-detailed depth estimation, and integrate Structure-Aware Enhancement Blocks at multiple scales to enhance radar features.
\end{itemize}
\section{RELATED WORK}
\label{sec:related work}

\subsection{Monocular Depth Estimation}
Monocular depth estimation aims to determine the depth of each pixel from an RGB image captured by a monocular camera. The development of deep learning has significantly advanced this field by facilitating the learning of depth features from some well-annotated datasets \cite{Geiger_Lenz_Stiller_Urtasun_2013,silberman2012indoor}. Eigen \textit{et al.} \cite{eigen2014depth} first introduce a multi-scale fusion network for depth regression. Following this, subsequent improvements have come from reinterpreting the regression task as a classification problem \cite{bhat2021adabins,Li_Wang_Liu_Jiang_2022}, incorporating additional priors \cite{shao2023nddepth,yang2023gedepth}, and developing more effective objective function \cite{xian2020structure,Yin_Liu_Shen_Yan_2019}. Despite these advances, generalizing to unseen domains remains a challenge. Recently, several methods have employed affine-invariant loss to enable multi-dataset joint training \cite{MiDaS,ZeroDepth,guizilini2023towards,Dany}. Among them, Depth Anything \cite{Dany} has shown leading performance in zero-shot monocular depth estimation. While it struggles to estimate accurate metric depth due to the lack of explicit depth cues, it excels at extracting structural information from unseen images, producing structure-detailed monocular depth.

\begin{figure*}[t]
      \centering
      \includegraphics[width=0.95\textwidth]{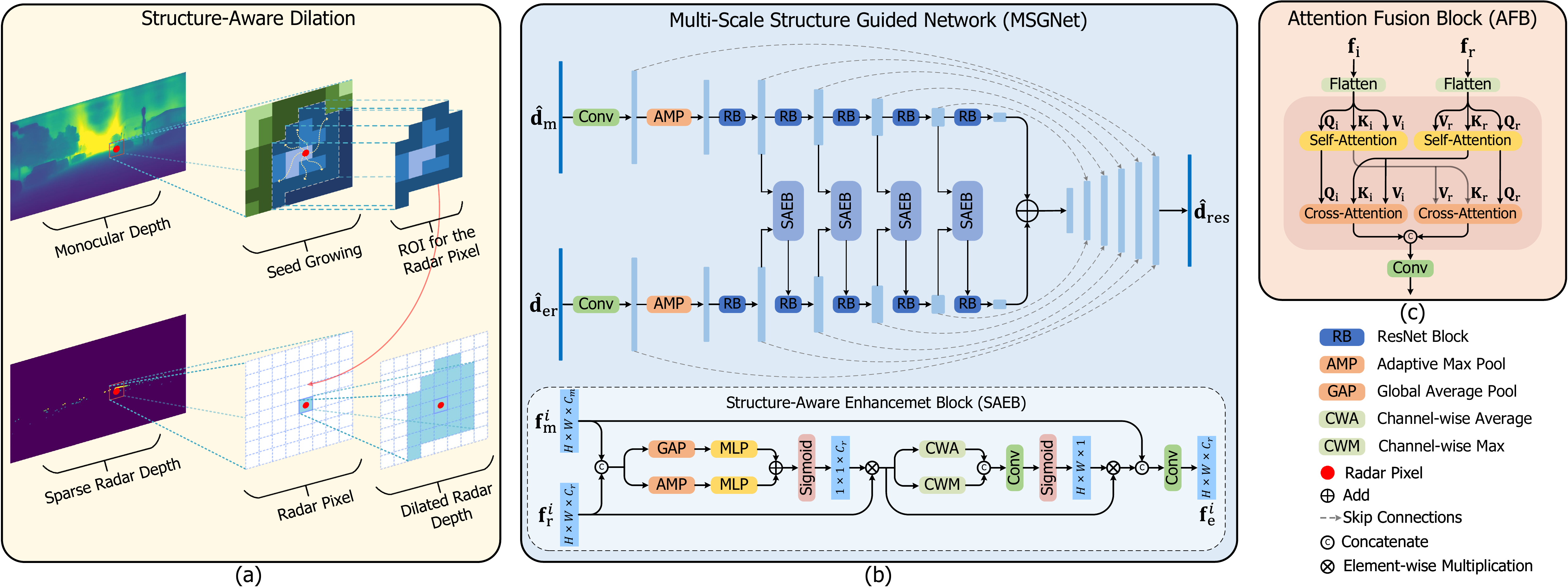}
      \caption{Detailed architectures of components in SA-RCD. (a) Diagram of structure-aware dilation. (b) Detailed structure of Multi-Scale Structure Guided Network (MSGNet) and Structure-Aware Enhancement Block (SAEB). (c) Detailed structure of Attention Fusion Block (AFB) in Radar-Camera Association Network (RCANet).}
      \label{fig:detailed structure}
      \vspace{-15pt} 
\end{figure*}

\subsection{Radar-Camera Depth Estimation}
Radar-camera depth estimation combines RGB images and radar depth to estimate dense metric depth. Due to the inherent noise and sparsity of radar data, enhancement operations are commonly applied before depth estimation. Lin \textit{et al.} \cite{LinDepth} reduces noise by predicting a coarse dense depth map, while Lo \textit{et al.} \cite{RegressionDepth} improves upon \cite{LinDepth} by introducing a densification operation that extends radar depth along the height dimension using fixed heights. Gasperini \textit{et al.} \cite{R4Dyn} further densifies depth by duplicating radar points at specific height intervals within object detection boxes, followed by bilateral filtering to remove points near object boundaries. In sight of the uncertainty of radar points, Long \textit{et al.} \cite{RC-PDA} propose a learning-based approach to determine radar-camera pixel association using multiple radar frames and RGB images for radar depth enhancement. Building on this, Singh \textit{et al.} \cite{Radarnet} and Li \textit{et al.} \cite{li2024radarcam} employ RadarNet to model the uncertain associations between predefined patches and radar points, producing semi-dense depth. Despite these advancements , current methods either rely on multiple radar frames or constrain ROIs for radar points to rigid rectangular shapes, failing to leverage structure priors in the scene. As a result, they may introduce errors and confusions in the enhanced radar data, which impedes accurate and structure-detailed metric depth estimation in the next stage.

\section{METHOD}
\label{sec:methodology}
Fig. \ref{fig:framework} illustrates the overall framework of SA-RCD, which includes three stages: monocular depth estimation, structure-aware radar depth enhancement, and metric depth estimation. Given an RGB image $\mathbf{I} \in \mathbb{R}^{3\times H\times W}$ and a single frame radar-projected depth map $\mathbf{d}_\text{r} \in \mathbb{R}^{H \times W}$, we aim to obtain a dense metric depth map $\mathbf{\hat{d}} \in \mathbb{R}^{H \times W}$. First, we leverage monocular depth estimation to extract structure priors from $\mathbf{I}$, generating a scaleless \cite{li2024radarcam} but structure-detailed monocular depth map $\hat{\mathbf{d}}_\text{m} \in \mathbb{R}^{H\times W}$. Then, in the structure-aware radar depth enhancement, we apply structure-aware dilation to $\mathbf{d}_\text{r}$, guided by the depth distribution of $\hat{\mathbf{d}}_\text{m}$, resulting in targeted ROIs for radar pixels and a dilated radar depth $\hat{\mathbf{d}}_\text{dr} \in \mathbb{R}^{H\times W}$. Next, we feed $\hat{\mathbf{d}}_\text{dr}$ into a Radar-Camera Association Network (RCANet) to learn the radar-camera pixel association within the ROIs, producing a confidence map $\hat{\mathbf{c}} \in \mathbf{[0,1]}^{H\times W}$. Subsequently, we use $\hat{\mathbf{c}}$ to filter out noise from $\hat{\mathbf{d}}_\text{dr}$, yielding the final enhanced radar depth map $\hat{\mathbf{d}}_\text{er} \in \mathbb{R}^{2\times H\times W}$. Finally, we combines  $\hat{\mathbf{d}}_\text{er}$ with $\hat{\mathbf{d}}_\text{m}$ by a Multi-scale Structure Guided Network (MSGNet) to estimate the dense metric depth $\hat{\mathbf{d}}$.

\subsection{Monocular Depth Prediction}
\label{Monocular Depth Prediction}
Although RGB images contain abundant structure priors, their direct applications in guiding the generation of ROIs for radar points are hindered by the inherent absence of geometric information and the interference of irrelevant textures. Considering this, we utilize a MDE model to effectively extract useful structure information from the given image, and generate a structure-detailed monocular depth map with fine relative geometric relationships. In this work, we choose the leading MDE model Depth Anything \cite{Dany} to predict monocular depth map $\hat{\mathbf{d}}_\text{m}$ from the input RGB image $\mathbf{I}$. Then we employ $\hat{\mathbf{d}}_\text{m}$ as an effective guidance to generate ROIs for radar points, thereby facilitating the enhancement of radar depth. Furthermore, we also leverage $\hat{\mathbf{d}}_\text{m}$ to enhance the radar features during the metric depth estimation through MSGNet.

\subsection{Structure-Aware Radar Depth Enhancement}
\label{Radar Depth Enhancement}
To address the inherent sparsity and noise of radar data, we perform structure-aware radar depth enhancement to enhance radar depth within a more targeted region. First, we develop a structure-aware dilation strategy to generate ROIs for all radar pixels in a seed-growing manner. Next, within these ROIs, we extend the depth of radar pixels for densification and train a Radar-Camera Association Network (RCANet) for confidence-based noise filtering.  

\textbf{Structure-Aware Dilation.} Given a radar pixel $\mathbf(i,j)$ with a depth value of $\mathbf{d}_\text{r}(i,j)$, we search for its region of interest (ROI), which is a spatially continuous set of pixels that are likely to share similar depth values with $\mathbf{d}_\text{r}(i,j)$. To achieve this, we develop a structure-aware dilation strategy by leveraging $\hat{\mathbf{d}}_\text{m}$ as the guidance, as illustrated in Fig. \ref{fig:detailed structure}{(a)}. This strategy is based on the assumption that pixels with similar radar depth values in $\mathbf{d}_\text{r}$ tend to share similar depth distribution in $\hat{\mathbf{d}}_\text{m}$. During the dilation process, the structure-related connectivity in  $\hat{\mathbf{d}}_\text{m}$ servers as an effective guidance to determine pixels that should be included in the ROI. Specially, for a radar depth $\mathbf{d}_\text{r}(i,j)$, we first set its corresponding monocular depth $\hat{\mathbf{d}}_\text{m}(i,j)$ as the seed point. Then, we dilate the ROI of radar pixel $\mathbf(i,j)$ outward from $\hat{\mathbf{d}}_\text{m}(i,j)$ in a seed-growing manner \cite{adams1994seeded}, until the depth differences between the candidate pixels and $\hat{\mathbf{d}}_\text{m}(i,j)$ exceed a specified tolerance \(\tau_1\):
\begin{align}
\mathcal{R}_{\mathbf(i,j)} = \{ (u, v) \mid |\hat{\mathbf{d}}_\text{m}(u, v) - \hat{\mathbf{d}}_\text{m}(i,j)| < \tau_{1} \}.
\end{align}
Here, $\mathcal{R}_{\mathbf(i,j)}$ represents the dilated ROI for $\mathbf{d}_\text{r}(i,j)$. $\tau_{1}$ controls the dilation process by ensuring that only pixels with depth values consistent with $\mathbf{d}_\text{m}(i,j)$ are included. In this way, we adapt the ROIs of radar pixels in $\mathbf{d}_\text{r}$ to the  depth distribution of $\hat{\mathbf{d}}_\text{m}$, producing more targeted range than a fixed or predefined area. Once $\mathcal{R}_{\mathbf(i,j)}$ is established, we extend the depth value of $\mathbf{d}_\text{r}(i,j)$ across it to generate a denser dilated radar depth of $\mathbf{d}_\text{r}(i,j)$:
\begin{align}
\hat{\mathbf{d}}_\text{dr}^{(i,j)}(u, v) = 
\begin{cases} 
\mathbf{d}_\text{r}(i,j) & \text{if } (u, v) \in \mathcal{R}_{\mathbf(i,j)} \\
0 & \text{otherwise}
\end{cases}.
\end{align}
By iterating the above process for all radar pixels, we continuously update the dilated radar depth and the ROIs for radar pixels. Ultimately, we yield a dilated radar depth map $\hat{\mathbf{d}}_\text{dr} \in \mathbb{R}^{H\times W}$ and a combined ROI for all radar pixels in $\mathbf{d}_\text{r}$:
\begin{align}
\mathcal{R} = \bigcup_{(i,j) \in \mathcal{P}} \mathcal{R}_{\mathbf(i,j)},
\end{align}
where $\mathcal{P}$ is the set of all radar pixels.

\textbf{Radar-Camera Pixel Association.} Although $\mathcal{R}$ is produced through structure-aware dilation, it remains suboptimal due to the complex uncertainty of radar points, which results in noise in $\hat{\mathbf{d}}_\text{dr}$. Considering this, we employ a U-Net based network \cite{unet}, named Radar-Camera Association Network (RCANet), to estimate radar-camera pixel association for noise filtering. Given the image $\mathbf{I}$ and the dilated radar depth map $\hat{\mathbf{d}}_\text{dr}$, we first extract features from them using a ResNet-34 backbone and a ResNet-18 backbone \cite{ResNet}, respectively. Following this, we feed the extracted image feature $\mathbf{f}_\text{i}$ and the radar feature $\mathbf{f}_\text{r}$  into a specialized attention-based fusion block (AFB) for cross-modal feature fusion, as illustrated in Fig. \ref{fig:detailed structure}{(c)}. AFB stacks $\mathit{N}$ modules combining self-attention and cross-attention, which are utilized for feature enhancement and feature interaction across modalities. After feature fusion through AFB, we input the fused features, along with the multi-scale features from both branches, to a U-Net decoder \cite{unet} with skip connections. Instead of estimating a confidence patch with a predefined shape for each radar pixel individually  \cite{Radarnet}, the decoder outputs a single confidence map $\hat{\mathbf{c}} \in \mathbf{[0,1]}^{H\times W}$, which contains valid values only within $\mathcal{R}$, focusing on a more targeted region. The value of each valid pixel represents the probability of its association with a radar pixel. For training RCANet, we generate the interpolated dense LiDAR depth map ${\mathbf{d}}_\text{int}$ via multi-frame fusion and interpolation (see Section \ref{dataset and metrics}). Then, we compute the ground truth confidence map ${\mathbf{c}}$ by comparing the absolute difference between $\hat{\mathbf{d}}_\text{dr}$ and ${\mathbf{d}}_\text{int}$ within $\mathcal{R}$:%
\begin{align}
\mathbf{c}(u, v) = 
\begin{cases} 
1 & \text{if } |\mathbf{d}_\text{int}(u, v) - \hat{\mathbf{d}}_\text{dr}(u, v)| \leq \tau_{2}\\
0 & \text{otherwise.}
\end{cases},
\end{align}
where ${(u, v)} \in \mathcal{R}$ and $\tau_{2}$ is a depth difference threshold. During the training process, we minimize a binary cross-entropy loss between $\hat{\mathbf{c}}$ and ${\mathbf{c}}$:
\begin{align}
\begin{aligned}\mathcal{L}_\text{conf}= & -\frac{1}{\left|\mathcal{R}\right|} \sum_{(u,v) \in \mathcal{R}}(\mathbf{c}(u,v) \log (\hat{\mathbf{c}}(u,v)) \\& +(1-\mathbf{c}(u,v)) \log (1-\hat{\mathbf{c}}(u,v))\end{aligned}.
\end{align}
%

\textbf{Confidence-based Noise Filtering.} After estimating radar-camera pixel association within $\mathcal{R}$ through RCANet, we filter out unreliable pixels in $\hat{\mathbf{d}}_\text{dr}$. Specifically, by referencing $\hat{\mathbf{c}}$, we consider pixels in $\hat{\mathbf{d}}_\text{dr}$ with lower confidence than a threshold $\tau_{3}$ as noise and remove them. In this way, we generate a radar depth map $\hat{\mathbf{d}}_\text{fr}\in \mathbb{R}^{H \times W}$, which is further enhanced through noise filtering after densification.
Then we concatenate $\mathbf{d}_\text{r}$ and $\hat{\mathbf{d}}_\text{fr}$ along the channel dimension to obtain the final enhanced radar depth map $\hat{\mathbf{d}}_\text{er} \in \mathbb{R}^{2\times H\times W}$.
\subsection{Metric Depth Estimation} 
\label{Monocular Depth Alignment}
After radar enhancement, we conduct metric depth estimation through a Multi-scale Structure Guided Network (MSGNet). Considering the detailed structures in $\hat{\mathbf{d}}_\text{m}$ and its proximity to the true depth distribution, we learn a residual map $\hat{\mathbf{d}}_\text{res}$ between $\hat{\mathbf{d}}_\text{m}$ and the ground truth depth. To make full use of the depth cues in  $\hat{\mathbf{d}}_\text{er}$ during this process, we design a Structure-Aware Enhancement Block (SAEB) to enhance the radar features. 
 \begin{figure*}[t]
\centering
\includegraphics[width=\textwidth]{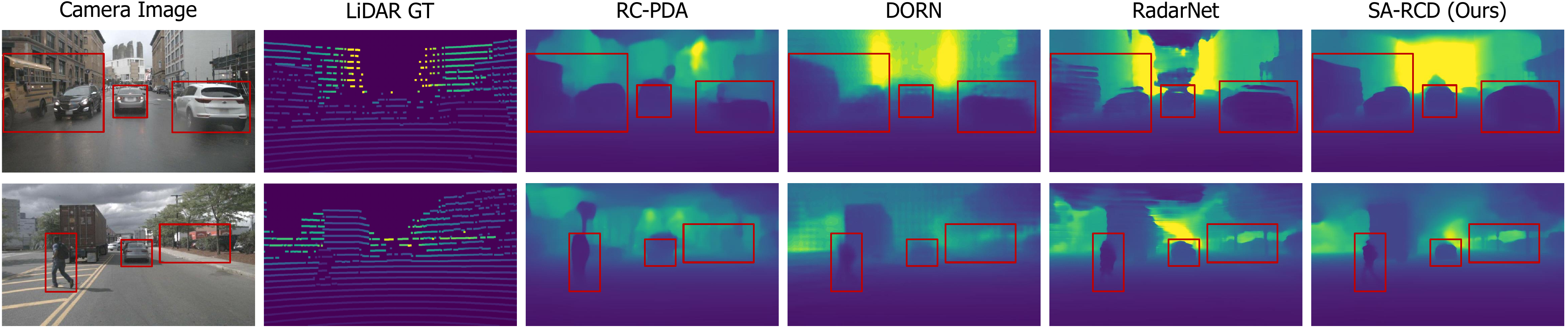}
   \caption{Qualitative comparison on the nuScenes test set at an 80-meter depth range. From left to right: RGB images from monocular camera, ground truth depth maps from LiDAR, dense metric depth maps estimated by RC-PDA \cite{RC-PDA}, DORN \cite{LinDepth}, RadarNet \cite{Radarnet}, and SA-RCD(Ours).}
\label{fig:qualitative comparisons}
\vspace{-5pt}
\end{figure*}

\textbf{Multi-scale Structure Guided Network (MSGNet).}
MSGNet takes $\hat{\mathbf{d}}_\text{m}$ and $\hat{\mathbf{d}}_\text{er}$ as inputs, extracting features from them through a ResNet-34 backbone and a ResNet-18 backbone, respectively. During feature extraction, we integrate the Structure-Aware Enhancement Block (SAEB) to guide the enhancement of radar features at multiple scales, as shown in Fig. \ref{fig:detailed structure}{(b)}. Then we add the features extracted from the two branches and feed the added feature to a multi-scale decoder with skip connections, which produces a residual map \(\hat{\mathbf{d}}_\text{res}\). We obtain the final predicted dense metric depth through \(\hat{\mathbf{d}} = \hat{\mathbf{d}}_\text{m} + \hat{\mathbf{d}}_\text{res}\). During the training of MSGNet, we accumulate adjacent LiDAR frames to generate the accumulated LiDAR depth map \(\mathbf{d}_\text{acc}\) (see Section \ref{dataset and metrics}). We employ a $L_{1}$ penalty to minimize the difference between $\hat{\mathbf{d}}$, and $\mathbf{d}_\text{acc}$ and $\mathbf{d}_\text{int}$:
\begin{align}
\mathcal{L}_\text{depth} &=  \frac{1}{|\Omega_\text{acc}|} \sum_{(u,v) \in \Omega_\text{acc}} |\mathbf{d}_\text{acc}(u,v) - \hat{\mathbf{d}}(u,v)| + \notag \\
&\quad \frac{\lambda}{|\Omega_\text{int}|} \sum_{(u,v) \in \Omega_\text{int}} |\mathbf{d}_\text{int}(u,v) - \hat{\mathbf{d}}(u,v)|.
\end{align}
Here,  \(\Omega_\text{acc}, \Omega_\text{int} \subset \Omega\) denote the set of pixels where $\mathbf{d}_\text{acc}$ and $\mathbf{d}_\text{int}$  have valid values, respectively. \(\lambda\) is a coefficient used to balance the two loss terms.

\textbf{Structure-Aware Enhancement Block (SAEB).} Although $\hat{\mathbf{d}}_\text{er}$ is denser and cleaner than ${\mathbf{d}}_\text{r}$, it remains sparse with poor structure information, leading to convolutions over many zero activations during radar feature encoding. To address this, we propose the Structure-Aware Enhancement Block (SAEB) to enhance multi-scale radar features, as shown in Fig. \ref{fig:detailed structure}{(b)}. Specifically, at the $i$-th scale, denote the monocular depth feature map as $\mathbf{f}_\text{m}^{i}\in \mathbb{R}^{h \times w \times c_{m}}$ and the radar feature map as $\mathbf{f}_\text{r}^{i}\in \mathbb{R}^{h \times w \times c_{r}}$, the operations in SAEB can be summarized as:
\begin{align}
\mathbf{f}_\text{c}^{i} = \mathbf{M}_\text{c}^{i}(\mathtt{Concat}(\mathbf{f}_\text{m}^{i},\mathbf{f}_\text{r}^{i})) \odot \mathbf{f}_\text{r}^{i},
\end{align}
\begin{align}
\mathbf{f}_\text{e}^{i} = \mathtt{Conv}(\mathtt{Concat}((\mathbf{M}_\text{s}^{i}(\mathbf{f}_\text{c}^{i}) \odot \mathbf{f}_\text{r}^{i}) ,\mathbf{f}_\text{m}^{i})).
\end{align}
Here, $\odot$ denotes element-wise multiplication. $\mathbf{M}_\text{c}^{i} \in \mathbb{R}^{1 \times 1 \times c_\text{r}}$ and $\mathbf{M}_\text{s}^{i} \in \mathbb{R}^{h \times w \times c_\text{r}}$ are the attention maps derived via channel attention and spatial attention operations \cite{woo2018cbam}. $\mathbf{f}_\text{c}^{i}$ is the intermediate channel-wise enhanced radar feature, while $\mathbf{f}_\text{e}^{i}$ is the final enhanced dense radar feature. $\mathtt{Conv}$ denotes a $1\times1$ convolution to match the feature dimensions with  $\mathbf{f}_\text{c}^{i}$. In this manner, SAEB captures high-frequency responses, which are typically indicative of structure information, from both the channel and spatial dimensions. It guides the radar features to focus on the channels and spatial locations that convey structure information.
\begin{figure}[t]
     \centering 
     \begin{subfigure}[t]{\linewidth}
      \centering
    \includegraphics[width=\linewidth]{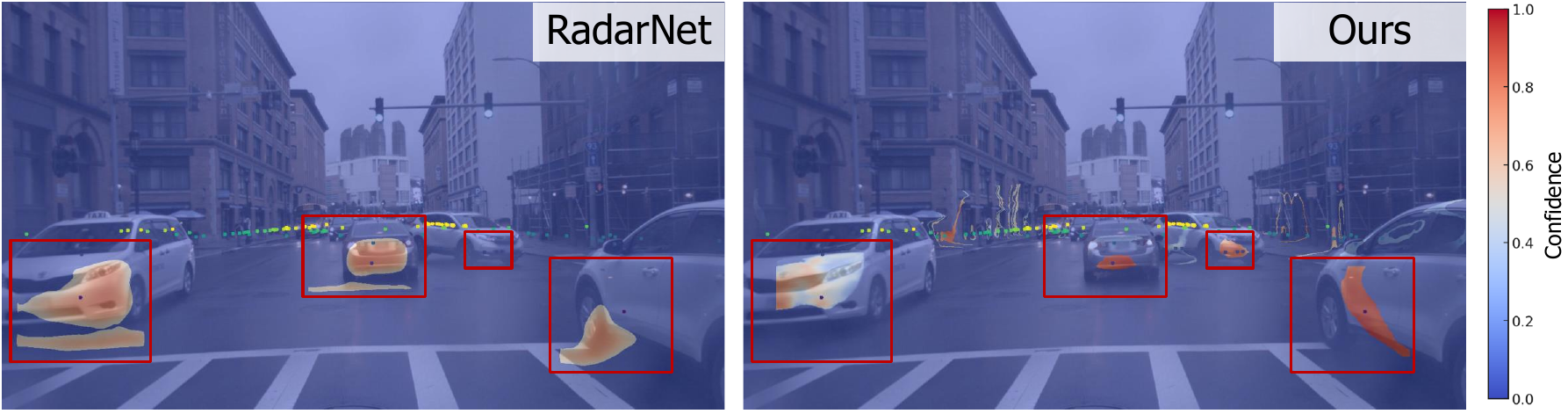}
    \vspace{-18pt}
    \captionsetup{font=footnotesize}
    \caption{An example of the confidence map.}
    \label{fig:head1}
     \end{subfigure}
     \begin{subfigure}[t]{\linewidth}
      \centering
    \includegraphics[width=\linewidth]{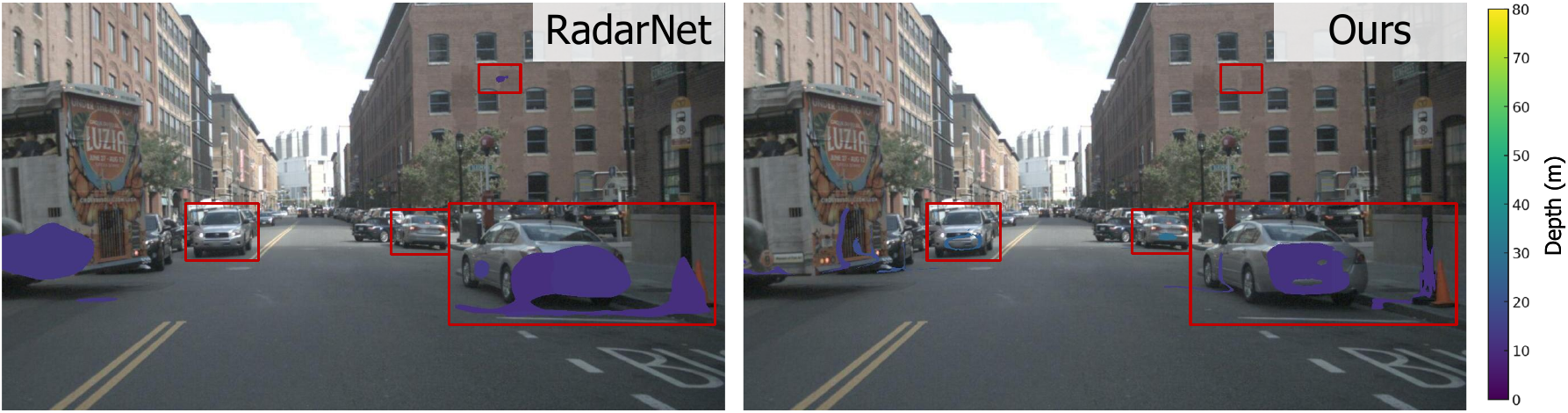}
    \vspace{-18pt}
    \captionsetup{font=footnotesize}
    \caption{An example of  the enhanced radar depth.}
    \label{fig:head2}
     \end{subfigure}
      \vspace{-5pt}
     \captionsetup{font=footnotesize}
     \caption{Qualitative comparison of the confidence maps and the enhanced radar depth maps between RadarNet \cite{Radarnet} and ours. In (a), each pixel of the confidence map represents the probability of its association with a radar pixel, with the radar depth visualized using points of different colors.  }
     \label{fig:qualitative comparisons of conf maps and enhanced radar} 
     \vspace{-17pt} 
\end{figure} 
\section{EXPERIMENTS}
\label{sec:experiement}
\subsection{Dataset and Metrics}
\label{dataset and metrics}
\textbf{Dataset.} The nuScenes benchmark \cite{caesar2020nuscenes} is a large-scale autonomous driving dataset collected in Boston and Singapore across diverse locations and weather conditions, providing data from LiDAR, mmWave radar, camera and IMU. It contains 1000 scenes of 20s duration each. Following the nuScenes train-test split, we employ 700 scenes for training, 150 scenes for validation, and 150 scenes for testing.

\textbf{Data Preprocessing.} Following ~\cite{Radarnet}, we accumulate LiDAR frames from both future and past time steps by projecting the point clouds onto the current frame to generate ${\mathbf{d}}_\text{acc}$. During this process, we remove the point clouds of dynamic objects from both future and past frames using the provided bounding boxes before projecting them onto to the current frame. Furthermore, we employ scaffolding \cite{wong2020unsupervised} to interpolate ${\mathbf{d}}_\text{acc}$, generating a dense LiDAR depth map $\mathbf{d}_\text{int}$. It should be noted that ${\mathbf{d}}_\text{acc}$ and $\mathbf{d}_\text{int}$ are only used during training, while the sparse LiDAR depth map, which is projected from single-frame LiDAR point cloud, is used as ground truth during evaluation.

\textbf{Metrics.} In correspondence with existing methods~\cite{Radarnet,LinDepth,RegressionDepth,RC-PDA,RCDPT}, we adopt mean absolute error (MAE) and root mean squared error (RMSE) for quantitative comparison. 
\begin{table}[th]
\centering
\captionsetup{justification=centering}
\caption{\textsc{Quantitative comparison on the nuScenes dataset. DORN utilizes 5 frames from 3 different radars to densify radar data. RC-PDA uses 3 images and 5 radar frames to compute "Flow". Results marked with \textsuperscript{\textdagger} are trained with sparse LiDAR supervision. All metrics are in millimeters.}}
\label{quantitative comparisons on nuscenes}
\resizebox{\linewidth}{!}{
\begin{tabular}{c|c|cc|cc}
\toprule
\textbf{Distance}& \textbf{Method} & \textbf{\# Radar Frames} & \textbf{\# Images} & \textbf{MAE$\downarrow$} & \textbf{RMSE$\downarrow$} \\
\hline
\multirow{4}*{0-50m}& RC-PDA \cite{RC-PDA} & 5 & 3 & 2225.0 & 4156.5 \\
{} &  RC-PDA + HG \cite{RC-PDA} & 5 & 3 & 2315.7 & 4321.6 \\
{} &  DORN \cite{RegressionDepth} & 5(x3) & 1 & 1926.6 & 4124.8 \\
{} &  RadarNet \cite{Radarnet}& 1 & 1  & 1727.7  & 3746.8 \\
{} &  \textbf{SA-RCD (Ours)}  & 1 & 1  & \textbf{1680.1}& \textbf{3735.4}\\
\hline
\multirow{4}*{0-70m}& RC-PDA \cite{RC-PDA} & 5 & 3 & 3326.1 & 6700.6 \\
{} &  RC-PDA + HG \cite{RC-PDA} & 5 & 3 & 3485.6 & 7002.9 \\
{} &  DORN \cite{RegressionDepth} & 5(x3) & 1 & 2380.6 & 5252.7 \\
{} &  RadarNet  \cite{Radarnet}& 1 & 1  & 2073.2  & 4590.7 \\
{} &  \textbf{SA-RCD (Ours)}  & 1 & 1  & \textbf{1990.3}& \textbf{4477.5}\\
\hline
\multirow{11}*{0-80m}& Depth Anything \cite{Dany} & 0 & 1 & 3558.6 & 6875.0 \\
{} & Sparse-to-dense \cite{Sparse-to-dense} & 3 & 1 & 2374.0 & 5628.0 \\
{} &  RC-PDA \cite{RC-PDA} & 5 & 3 & 3713.6 & 7692.8 \\
{} &  RC-PDA with HG \cite{RC-PDA} & 5 & 3 & 3884.3 & 8008.6 \\
{} &  DORN \cite{RegressionDepth} & 5(x3) & 1 & 2467.7 & 5554.3 \\
{} &  Lin \textit{et al.} \cite{LinDepth}& 3 & 1 & 2371.0 & 5623.0 \\
{} &  R4Dyn \cite{R4Dyn} & 4 & 1  & N/A & 6434.0 \\
{} &  RadarNet \cite{Radarnet}& 1 & 1  & 2179.3 & 4898.7 \\
{} &  \textbf{SA-RCD (Ours)}  & 1 & 1  & \textbf{2082.9}& \textbf{4747.8}\\
\cline{2-6}
{} &  RadarCam-Depth\textsuperscript{\textdagger} \cite{li2024radarcam} & 1 & 1 & 1689.7 & 3948.0  \\
{} &  Sparse Beats Dense\textsuperscript{\textdagger} \cite{li2023sparse} & 1 & 1  & 1927.0 & 4609.6 \\
{} &  \textbf{SA-RCD (Ours)\textsuperscript{\textdagger}}  & 1 & 1  & \textbf{1490.3}& \textbf{3610.8}\\
\bottomrule
\end{tabular}
}
\vspace{-10pt} 
\end{table}

\subsection{Implementation Details}
For RCANet, we train it for 200 epoches with a batch size of 6. The input image and radar depth map are randomly cropped into patches of $352\times 704$. The learning rate for RCANet is set to $3\times10^{-4}$. For MSGNet, we train it for 300 epochs with a batch size of 16. The inputs are randomly cropped into patches of $448 \times 448$. The learning rate for MSGNet is initially set to $1\times10^{-3}$, then decreased to $5\times10^{-4}$ by 200 epochs, and finally decreased to $1\times10^{-4}$ after an additional 50 epochs. During the training process of both RCANet and MSGNet, we employ Adam Optimizer with $\beta_1=0.9$, $\beta_2=0.999$. Additionally, we apply data augmentations including horizontal flipping and adjustments to saturation, brightness, and contrast. For the hyperparameters mentioned in this paper, we empirically set $\tau_{1}=0.2~\mathrm{m}$, $\tau_{2}=0.4~\mathrm{m}$, $\tau_{3}=0.5$, and $\lambda=2$. All the experiments are conducted on an NVIDIA RTX 3090 GPU.
\begin{table*}[th]
	\centering
	\renewcommand\arraystretch{1.1}
 \captionsetup{justification=centering}
 	\caption{\textsc{Ablation Study. Mono Depth: monocular depth. Depth Map: learning a complete dense metric depth map from scratch. Extension, Filtering: radar depth extension and noise filtering through structure-aware radar depth enhancement.}}
	\resizebox{0.88\textwidth}{!}{
		\begin{tabular}{c|ccc|cc|ccc|ccc|cc}
			\toprule
			\multirow{2}{*}{\textbf{Settings}}&  \multicolumn{3}{c|}{\textbf{Inputs}} & \multicolumn{2}{c|}{\textbf{Target output}} & \multicolumn{3}{c|}{\textbf{Radar Enhancement}} & \multicolumn{3}{c|}{\textbf{Guidance Block}} & \multirow{2}{*}{\textbf{MAE$\downarrow$}} & \multirow{2}{*}{\textbf{RMSE$\downarrow$}} \\ \cline{2-12}
			           & \textbf{Radar} & \textbf{Image}     & \textbf{Mono Depth}& \textbf{Depth Map}& \textbf{Residual}   & \textbf{Extention}& \textbf{Filtering}& \textbf{AFB}        & \textbf{Add}        & \textbf{Concat}     & \textbf{SAEB}   & & \\
			\hline
			baseline & \checkmark & \checkmark &            & \checkmark &            &            &            & 		   & 		    &            &            & 2678.9  & 5246.4  \\ 
			  (a)      & \checkmark &            & \checkmark & \checkmark &            &            &            &            &            &            &            & 2321.8  & 4912.8  \\
			(b)      & \checkmark &            & \checkmark &            & \checkmark &            &            &            &            &            &            & 2196.9  & 4880.4  \\
            \hline
                (c)      & \checkmark &            & \checkmark &            & \checkmark & \checkmark &            &            &            &            &            & 2203.2  & 4822.5  \\
			(d)      & \checkmark &            & \checkmark &            & \checkmark & \checkmark & \checkmark &            &            &            &            & 2158.8  & 4798.7  \\
			(e)      & \checkmark &            & \checkmark &            & \checkmark & \checkmark & \checkmark & \checkmark &            &            &            & 2120.7  & 4754.8  \\
            \hline
			(f)      & \checkmark &            & \checkmark &            & \checkmark & \checkmark & \checkmark & \checkmark & \checkmark &            &            & 2131.3  & 4774.6  \\
			(g)      & \checkmark &            & \checkmark &            & \checkmark & \checkmark & \checkmark & \checkmark &            & \checkmark &            & 2189.8  & 4795.8  \\ 
			(h)      & \checkmark &            & \checkmark &            & \checkmark & \checkmark & \checkmark & \checkmark &            &            & \checkmark & \textbf{2082.9} & \textbf{4747.8} \\ 
			\bottomrule
		\end{tabular}
	}
	\label{ablation study table}
	\vspace{-2mm}
\end{table*}

\begin{table}[th]
    \centering
    \renewcommand\arraystretch{1.1}
    \captionsetup{justification=centering}
    \caption{\textsc{Comparison of Inference Time (s).}}
    \resizebox{0.96\linewidth}{!}{
        \begin{tabular}{c|ccc}
            \toprule
            \multirow{1}{*}{\textbf{Method}} & \multicolumn{1}{c}{\textbf{Depth Enhancement}} & \multicolumn{1}{c}{\textbf{Depth Estimation}} &\multicolumn{1}{c}{\textbf{Full Model}}\\ 
            \cline{1-4}
            
            \multirow{1}{*}{RadarNet\cite{Radarnet}}
            & \multicolumn{1}{c}{0.0994} & \multicolumn{1}{c}{0.0165} & 0.1159  \\

            \multirow{1}{*}{RC-PDA\cite{RC-PDA}}
            & \multicolumn{1}{c}{0.7111} & \multicolumn{1}{c}{0.0116} & 0.7227  \\

            \multirow{1}{*}{\textbf{SA-RCD (Ours)}} 
            &\textbf{0.5337} & \textbf{0.0196} & \textbf{0.5533}  \\

            \bottomrule
        \end{tabular}
    }
    \label{running time_table}
    \vspace{-4mm}
\end{table}

\subsection{Quantitative Results}
\label{quantitative results}
We evaluate our SA-RCD against existing radar-camera depth estimation methods \cite{Sparse-to-dense,RC-PDA,RegressionDepth,LinDepth,R4Dyn,Radarnet} on the nuScenes test set. Considering the usable range  \cite{Radarnet}, we compare performances at 50, 70, and 80-meter intervals, as shown in Table \ref{quantitative comparisons on nuscenes}. Across the three distance ranges, SA-RCD surpasses all compared methods in both MAE and RMSE. A recent study \cite{li2023sparse} observes that while sparse LiDAR supervision introduces stripe-like artifacts, it improves accuracy. Following this, we employ sparse LiDAR supervision for MSGNet and compared the metric with other competitive models trained in this manner. SA-RCD still shows a notable accuracy advantage over Sparse Beats Dense \cite{li2023sparse} and RadarCam-Depth \cite{li2024radarcam}.

\subsection{Qualitative Results}
\label{qualitative results}
Fig. \ref{fig:qualitative comparisons} shows two scenarios at distances of 80 meters for qualitative comparison. In the first row, SA-RCD captures fine structures and sharp depth edges for all four vehicles. In contrast, the results of RC-PDA, DORN, and RadarNet exhibit either incomplete structures or unclear depth edges. The second row shows a more challenging scene, including a pedestrian whose color is similar to the truck behind. SA-RCD is the only method that capture the complete structure of the pedestrian. In Fig. \ref{fig:qualitative comparisons of conf maps and enhanced radar}, we further evaluate the effectiveness of structure-aware radar depth enhancement by comparing our confidence map and enhanced radar depth map against those of RadarNet. Fig. \ref{fig:qualitative comparisons of conf maps and enhanced radar}{(a)} demonstrates that the high-confidence regions of our confidence maps are concentrated on cars, clearly separated from the ground. In Fig. \ref{fig:qualitative comparisons of conf maps and enhanced radar}{(b)}, the comparison of enhanced radar depth shows that RadarNet merges the depths of the car and the lamppost, whereas SA-RCD clearly distinguishes between them. 

\subsection{Ablation Study}
\label{ablation study}
To verify the effectiveness of components in SA-RCD, we conduct ablation studies over the 0-80m range, as shown in Table \ref{ablation study table}. The baseline can be considered as a simplified version of the FusionNet in \cite{Radarnet}, with the image encoder replaced by a ResNet-34 backbone. It uses two encoders to extract features from images and raw radar depth maps, and then learns a complete dense metric depth map from scratch. 

\textbf{Effect of  MDE.}
To demonstrate whether MDE is beneficial for radar depth estimation, we replace the input to the image branch of the baseline with the monocular depth map estimated by MDE in setting (a), leading to significant improvement of the metrics. Furthermore, in setting (b), we learn a residual map between the monocular depth map and ground truth, rather than learning a complete dense metric depth from scratch, which results in additional improvements.

\textbf{Effect of radar depth enhancement.}
To illustrate the effectiveness of structure-aware radar depth enhancement, we progressively incorporate steps and modules into the enhancement process. In setting (c), we extend radar depth with structure-aware dilation. Although RMSE is decreased, MAE is increased. This indicates that extension alone may amplify noise. Then, in setting (d), we filter the noise through RCANet, which demonstrates notable improvements. Furthermore, embedding AFB within RCANet in setting (e) leads to extra gains of the performance. In summary, the application of structure-aware radar enhancement brings notable improvements compared to using raw radar data.

\textbf{Effect of SAEB.}
We explore various feature fusion strategies to demonstrate the effectiveness of SAEB. In setting (f) and (g), we replace SAEB with two simpler fusion methods: add and concat. It can be observed that SAEB in setting (h) achieves better performance compared to add and concat.

\subsection{Comparison On Inference Time}
Table \ref{running time_table} shows the inference time of SA-RCD and several other methods during the radar depth enhancement stage, the depth estimation stage, and the full model, respectively. The inference time of SA-RCD is between that of RadarNet \cite{Radarnet} and RC-PDA \cite{RC-PDA}. In the radar depth enhancement stage of SA-RCD, the majority of the inference time is consumed by the structure-aware dilation, which takes 0.4919~$\mathrm{s}$.

\section{Conclusions}
\label{sec:conclusion}
In this work, we propose SA-RCD, a novel radar-camera depth estimation framework with three stages to estimate accurate and structure-detailed metric depth. In the first stage, we leverage the structure priors of RGB images through a powerful MDE model. In the second stage, we effectively address the sparsity and noise of radar depth through structure-aware radar depth enhancement. In the final stage, we achieve structure-detailed depth estimation using a well-designed MSGNet. We experimentally show that SA-RCD achieves state-of-the-art performance in both quantitative and qualitative evaluations on the  nuScenes dataset.

{
\AtNextBibliography{\small}
\printbibliography

@article{song2021self,
  title={Self-supervised depth completion from direct visual-lidar odometry in autonomous driving},
  author={Song, Zhenbo and Lu, Jianfeng and Yao, Yazhou and Zhang, Jian},
  journal={IEEE Transactions on Intelligent Transportation Systems},
  volume={23},
  number={8},
  year={2021},
  pages={11654--11665},
}

@inproceedings{li2023bevdepth,
  title={B{EVD}epth: Acquisition of reliable depth for multi-view 3{D} object detection},
  author={Li, Yinhao and Ge, Zheng and Yu, Guanyi and Yang, Jinrong and Wang, Zengran and Shi, Yukang and Sun, Jianjian and Li, Zeming},
  booktitle={Proceedings of the AAAI Conference on Artificial Intelligence},
  volume={37},
  number={2},
  pages={1477--1485},
  year={2023}
}

@article{han23064d,
  title={4{D} millimeter-wave radar in autonomous driving: A survey},
  author={Han, Z. and Wang, J. and Xu, Z. and Yang, S. and He, L. and Xu, S. and Wang, J.},
  journal={arXiv preprint arXiv:2306.04242},
  year={2023}
}

@article{cui2021deep,
  title={Deep learning for image and point cloud fusion in autonomous driving: A review},
  author={Cui, Yaodong and Chen, Ren and Chu, Wenbo and Chen, Long and Tian, Daxin and Li, Ying and Cao, Dongpu},
  journal={IEEE Transactions on Intelligent Transportation Systems},
  volume={23},
  number={2},
  pages={722--739},
  year={2021},
}

@article{zhou2022towards,
  title={Towards deep radar perception for autonomous driving: Datasets, methods, and challenges},
  author={Zhou, Yi and Liu, Lulu and Zhao, Haocheng and López-Benítez, Miguel and Yu, Limin and Yue, Yutao},
  journal={Sensors},
  volume={22},
  number={11},
  pages={4208},
  year={2022},
}

@article{alaba2024emerging,
  title={Emerging trends in autonomous vehicle perception: Multimodal fusion for 3{D} object detection},
  author={Alaba, Simegnew Yihunie and Gurbuz, Ali C and Ball, John E},
  journal={World Electric Vehicle Journal},
  volume={15},
  number={1},
  pages={20},
  year={2024},
}

@article{wang2023multi,
  title={Multi-modal 3{D} object detection in autonomous driving: A survey and taxonomy},
  author={Wang, Li and Zhang, Xinyu and Song, Ziying and Bi, Jiangfeng and Zhang, Guoxin and Wei, Haiyue and Tang, Liyao and Yang, Lei and Li, Jun and Jia, Caiyan and others},
  journal={IEEE Transactions on Intelligent Vehicles},
  volume={8},
  number={7},
  pages={3781--3798},
  year={2023},
}

@article{Zoedepth,
  title={Zoe{D}epth: Zero-shot transfer by combining relative and metric depth},
  author={Bhat, Shariq Farooq and Birkl, Reiner and Wofk, Diana and Wonka, Peter and Müller, Matthias},
  journal={arXiv preprint arXiv:2302.12288},
  year={2023}
}

@inproceedings{ZeroDepth,
  title={Towards zero-shot scale-aware monocular depth estimation},
  author={Guizilini, Vitor and Vasiljevic, Igor and Chen, Dian and Ambruș, Rareș and Gaidon, Adrien},
  booktitle={Proceedings of the IEEE/CVF International Conference on Computer Vision},
  pages={9233--9243},
  year={2023}
}

@inproceedings{Dany,
  title={Depth {A}nything: Unleashing the power of large-scale unlabeled data},
  author={Yang, Lihe and Kang, Bingyi and Huang, Zilong and Xu, Xiaogang and Feng, Jiashi and Zhao, Hengshuang},
  booktitle={Proceedings of the IEEE/CVF Conference on Computer Vision and Pattern Recognition},
  pages={10371--10381},
  year={2024}
}

@article{MiDaS,
  title={Towards robust monocular depth estimation: Mixing datasets for zero-shot cross-dataset transfer},
  author={Ranftl, René and Lasinger, Katrin and Hafner, David and Schindler, Konrad and Koltun, Vladlen},
  journal={IEEE Transactions on Pattern Analysis and Machine Intelligence},
  volume={44},
  number={3},
  pages={1623--1637},
  year={2020},
}

@inproceedings{long2021full,
  title={Full-velocity radar returns by radar-camera fusion},
  author={Long, Yunfei and Morris, Daniel and Liu, Xiaoming and Castro, Marcos and Chakravarty, Punarjay and Narayanan, Praveen},
  booktitle={Proceedings of the IEEE/CVF International Conference on Computer Vision},
  pages={16198--16207},
  year={2021}
}

@inproceedings{RC-PDA,
  title={Radar-camera pixel depth association for depth completion},
  author={Long, Yunfei and Morris, Daniel and Liu, Xiaoming and Castro, Marcos and Chakravarty, Punarjay and Narayanan, Praveen},
  booktitle={Proceedings of the IEEE/CVF Conference on Computer Vision and Pattern Recognition},
  pages={12507--12516},
  year={2021}
}

@inproceedings{Radarnet,
  title={Depth estimation from camera image and mmWave radar point cloud},
  author={Singh, Akash Deep and Ba, Yunhao and Sarker, Ankur and Zhang, Howard and Kadambi, Achuta and Soatto, Stefano and Srivastava, Mani and Wong, Alex},
  booktitle={Proceedings of the IEEE/CVF Conference on Computer Vision and Pattern Recognition},
  pages={9275--9285},
  year={2023}
}

@inproceedings{RCDPT,
  title={{RCDPT}: Radar-camera fusion dense prediction transformer},
  author={Lo, Chen-Chou and Vandewalle, Patrick},
  booktitle={Proceedings of the IEEE International Conference on Acoustics, Speech and Signal Processing},
  pages={1--5},
  year={2023}
}

@inproceedings{ResNet,
  title={Deep residual learning for image recognition},
  author={He, Kaiming and Zhang, Xiangyu and Ren, Shaoqing and Sun, Jian},
  booktitle={Proceedings of the IEEE/CVF Conference on Computer Vision and Pattern Recognition},
  pages={770--778},
  year={2016}
}

@article{wong2020unsupervised,
  title={Unsupervised depth completion from visual inertial odometry},
  author={Wong, Alex and Fei, Xiaohan and Tsuei, Stephanie and Soatto, Stefano},
  journal={IEEE Robotics and Automation Letters},
  volume={5},
  number={2},
  pages={1899--1906},
  year={2020}
}

@inproceedings{caesar2020nuscenes,
  title={Nu{S}cenes: A multimodal dataset for autonomous driving},
  author={Caesar, Holger and Bankiti, Varun and Lang, Alex H and Vora, Sourabh and Liong, Venice Erin and Xu, Qiang and Krishnan, Anush and Pan, Yu and Baldan, Giancarlo and Beijbom, Oscar},
  booktitle={Proceedings of the IEEE/CVF Conference on Computer Vision and Pattern Recognition},
  pages={11621--11631},
  year={2020}
}

@inproceedings{R4Dyn,
  title={R4{D}yn: Exploring radar for self-supervised monocular depth estimation of dynamic scenes},
  author={Gasperini, Stefano and Koch, Patrick and Dallabetta, Vinzenz and Navab, Nassir and Busam, Benjamin and Tombari, Federico},
  booktitle={Proceedings of the International Conference on 3{D} Vision},
  pages={751--760},
  year={2021}
}

@inproceedings{LinDepth,
  title={Depth estimation from monocular images and sparse radar data},
  author={Lin, Juan-Ting and Dai, Dengxin and Van Gool, Luc},
  booktitle={Proceedings of the International Conference on Intelligent Robots and Systems},
  pages={10233--10240},
  year={2020}
}

@inproceedings{RegressionDepth,
  title={Depth estimation from monocular images and sparse radar using deep ordinal regression network},
  author={Lo, Chen-Chou and Vandewalle, Patrick},
  booktitle={Proceedings of the IEEE International Conference on Image Processing},
  pages={3343--3347},
  year={2021}
}

@inproceedings{Sparse-to-dense,
  title={Sparse-to-dense: Depth prediction from sparse depth samples and a single image},
  author={Ma, Fangchang and Karaman, Sertac},
  booktitle={Proceedings of the IEEE International Conference on Robotics and Automation},
  pages={4796--4803},
  year={2018}
}

@inproceedings{guizilini2023towards,
  title={Towards zero-shot scale-aware monocular depth estimation},
  author={Guizilini, Vitor and Vasiljevic, Igor and Chen, Dian and Ambruș, Rareș and Gaidon, Adrien},
  booktitle={Proceedings of the IEEE/CVF International Conference on Computer Vision},
  pages={9233--9243},
  year={2023}
}

@inproceedings{unet,
  title={U-{N}et: Convolutional networks for biomedical image segmentation},
  author={Ronneberger, Olaf and Fischer, Philipp and Brox, Thomas},
  booktitle={Proceedings of the International Conference on Medical Image Computing and Computer-Assisted Intervention},
  pages={234--241},
  year={2015}
}

@inproceedings{yan2022rignet,
  title={Rig{N}et: Repetitive image guided network for depth completion},
  author={Yan, Zhiqiang and Wang, Kun and Li, Xiang and Zhang, Zhenyu and Li, Jun and Yang, Jian},
  booktitle={Proceedings of the European Conference on Computer Vision},
  pages={214--230},
  year={2022},
}

@inproceedings{eigen2014depth,
    author = {Eigen, David and Puhrsch, Christian and Fergus, Rob},
    title = {Depth map prediction from a single image using a multi-scale deep network},
    booktitle = {Proceedings of the International Conference on Neural Information Processing Systems},
    pages = {pp. 2366–2374},
    year = {2014},
}

@inproceedings{xian2020structure,
  title={Structure-guided ranking loss for single image depth prediction},
  author={Xian, Ke and Zhang, Jianming and Wang, Oliver and Mai, Long and Lin, Zhe and Cao, Zhiguo},
  booktitle={Proceedings of the IEEE/CVF Conference on Computer Vision and Pattern Recognition},
  pages={611--620},
  year={2020}
}

@article{adams1994seeded,
  title={Seeded region growing},
  author={Adams, Rolf and Bischof, Leanne},
  journal={IEEE Transactions on Pattern Analysis and Machine Intelligence},
  volume={16},
  number={6},
  pages={641--647},
  year={1994},
}

@inproceedings{woo2018cbam,
  title={{CBAM}: Convolutional block attention module},
  author={Woo, Sanghyun and Park, Jongchan and Lee, Joon-Young and Kweon, In So},
  booktitle={Proceedings of the European Conference on Computer Vision},
  pages={3--19},
  year={2018}
}

@article{CGFormer,
      title={Context and geometry aware voxel transformer for semantic scene completion}, 
      author={Zhu, Yu and Zhang, Runming and Ying, Jiacheng and Yu, Junchen and Hu, Xiaohai and Luo, Lun and Cao, Siyuan and Shen, Huiliang},
      journal={arXiv preprint arXiv:2405.13675},
      year={2024}
}

@article{Geiger_Lenz_Stiller_Urtasun_2013,
  title={Vision meets robotics: The {KITTI} dataset},
  author={Geiger, Andreas and Lenz, Philip and Stiller, Christoph and Urtasun, Raquel},
  journal={The International Journal of Robotics Research},
  volume={32},
  number={11},
  pages={1231--1237},
  year={2013}
}

@inproceedings{silberman2012indoor,
  title={Indoor segmentation and support inference from {RGBD} images},
  author={Silberman, Nathan and Hoiem, Derek and Kohli, Pushmeet and Fergus, Rob},
  booktitle={Proceedings of the European Conference on Computer Vision},
  pages={746--760},
  year={2012}
}

@inproceedings{bhat2021adabins,
  title={Ada{B}ins: Depth estimation using adaptive bins},
  author={Bhat, Shariq Farooq and Alhashim, Ibraheem and Wonka, Peter},
  booktitle={Proceedings of the IEEE/CVF Conference on Computer Vision and Pattern Recognition},
  pages={4009--4018},
  year={2021}
}

@ARTICLE{Li_Wang_Liu_Jiang_2022,
  author={Li, Zhenyu and Wang, Xuyang and Liu, Xianming and Jiang, Junjun},
  journal={IEEE Transactions on Image Processing}, 
  title={BinsFormer: Revisiting Adaptive Bins for Monocular Depth Estimation}, 
  year={2024},
  volume={33},
  pages={3964-3976}
}

@inproceedings{shao2023nddepth,
  title={N{DD}epth: Normal-distance assisted monocular depth estimation},
  author={Shao, Shuwei and Pei, Zhongcai and Chen, Weihai and Wu, Xingming and Li, Zhengguo},
  booktitle={Proceedings of the IEEE/CVF International Conference on Computer Vision},
  pages={7931--7940},
  year={2023}
}

@inproceedings{yang2023gedepth,
  title={G{ED}epth: Ground embedding for monocular depth estimation},
  author={Yang, Xiaodong and Ma, Zhuang and Ji, Zhiyu and Ren, Zhe},
  booktitle={Proceedings of the IEEE/CVF International Conference on Computer Vision},
  pages={12719--12727},
  year={2023}
}

@inproceedings{Yin_Liu_Shen_Yan_2019,
  title={Enforcing geometric constraints of virtual normal for depth prediction},
  author={Yin, Wei and Liu, Yifan and Shen, Chunhua and Yan, Youliang},
  booktitle={Proceedings of the IEEE/CVF International Conference on Computer Vision},
  pages={5684--5693},
  year={2019}
}

@inproceedings{tpvd,
  title={Tri-perspective view decomposition for geometry-aware depth completion},
  author={Yan, Zhiqiang and Lin, Yuankai and Wang, Kun and Zheng, Yupeng and Wang, Yufei and Zhang, Zhenyu and Li, Jun and Yang, Jian},
  booktitle={Proceedings of the IEEE/CVF Conference on Computer Vision and Pattern Recognition},
  pages={4874--4884},
  year={2024}
}

@inproceedings{pointdc,
  title={Aggregating feature point cloud for depth completion},
  author={Yu, Zhu and Sheng, Zehua and Zhou, Zili and Luo, Lun and Cao, Si-Yuan and Gu, Hong and Zhang, Huaqi and Shen, Hui-Liang},
  booktitle={Proceedings of the IEEE/CVF International Conference on Computer Vision},
  pages={8732--8743},
  year={2023}
}

@inproceedings{cspn,
  title={Depth estimation via affinity learned with convolutional spatial propagation network},
  author={Cheng, Xinjing and Wang, Peng and Yang, Ruigang},
  booktitle={Proceedings of the European Conference on Computer Vision},
  pages={103--119},
  year={2018}
}

@ARTICLE{cspn2,
  author={Cheng, Xinjing and Wang, Peng and Yang, Ruigang},
  journal={IEEE Transactions on Pattern Analysis and Machine Intelligence}, 
  title={Learning Depth with Convolutional Spatial Propagation Network}, 
  year={2020},
  volume={42},
  number={10},
  pages={2361-2379},
  doi={10.1109/TPAMI.2019.2947374}}

@inproceedings{nlspn,
  title={Non-local spatial propagation network for depth completion},
  author={Park, Jinsun and Joo, Kyungdon and Hu, Zhe and Liu, Chi-Kuei and Kweon, In So},
  booktitle={Proceedings of the European Conference on Computer Vision},
  pages={120--136},
  year={2020}
}

@article{li2024radarcam,
  title={RadarCam-{D}epth: Radar-camera fusion for depth estimation with learned metric scale},
  author={Li, Han and Ma, Yukai and Gu, Yaqing and Hu, Kewei and Liu, Yong and Zuo, Xingxing},
  journal={arXiv preprint arXiv:2401.04325},
  year={2024}
}

@article{li2023sparse,
  title={Sparse beats dense: Rethinking supervision in radar-camera depth completion},
  author={Li, Huadong and Jing, Minhao and Liang, Jiajun and Fan, Haoqiang and Ji, Renhe},
  journal={arXiv preprint arXiv:2312.00844},
  year={2023}
}
}

\end{document}